\documentclass[runningheads]{llncs}
\usepackage[T1]{fontenc}
\usepackage{graphicx}
\usepackage{amsmath}
\usepackage{amssymb}
\usepackage{amsfonts}
\usepackage{algorithm}
\usepackage{algorithmic}
\usepackage{subcaption}
\usepackage{cleveref}
\crefname{equation}{Eq.}{Eqs.}
\crefname{theorem}{Theorem}{Theorems}
\crefname{figure}{Fig.}{Fig.}
\crefname{section}{Section}{Sections}
\begin{document}
\title{Dynamic Feature Selection from Variable Feature Sets Using Features of Features}
\titlerunning{DFS from Variable Feature Sets Using Features of Features}
%
\author{Katsumi Takahashi\inst{1} \and
Koh Takeuchi\inst{1} \and
Hisashi Kashima\inst{1}}
\authorrunning{K. Takahashi, et al.}
%
\institute{Kyoto University, Kyoto, Japan}

\maketitle              
\begin{abstract}
Machine learning models usually assume that a set of feature values used to obtain an output is fixed in advance.
However, in many real-world problems, a cost is associated with measuring these features. 
To address the issue of reducing measurement costs, various methods have been proposed to dynamically select which features to measure, but existing methods assume that the set of measurable features remains constant, which makes them unsuitable for cases where the set of measurable features varies from instance to instance. 
To overcome this limitation, we define a new problem setting for Dynamic Feature Selection (DFS) with variable feature sets and propose a deep learning method that utilizes prior information about each feature, referred to as ''features of features''.
Experimental results on several datasets demonstrate that the proposed method effectively selects features based on the prior information, even when the set of measurable features changes from instance to instance.

\end{abstract}
\section{Introduction}
\label{intro}

In typical supervised machine learning, we input some or many feature values into the machine learning model and it makes a prediction about the label.
However, in many real-world problems, measuring these feature values themselves is costly and it is desirable to reduce the cost of measuring feature values.
For example, in medical diagnosis, the doctor performs some tests on the patient to distinguish the patient's disease, and it takes the financial cost or the patient's burden for these tests.
Therefore, it is better that the number of tests is smaller and the doctor sequentially selects a test to perform based on the results of the previous tests.
This problem is known as Dynamic Feature Selection (DFS) and has been addressed with various machine learning methods.

Compared to conventional static feature selection that selects only one feature set for one dataset \cite{static_feature_selection}, DFS selects a different feature set for each instance based on measured feature values.
This DFS problem typically has been addressed with reinforcement learning method \cite{datum-wise,RL_selection,joint_afa,opportunistic_learning}.
There are also DFS methods without reinforcement learning, which are based on information gain, variance of feature value, or mutual information \cite{matrix_completion_afa,cmi_feature,question_ordering}.
Notably, one of the latest methods \cite{cmi_feature} trains policy and predictor networks based on conditional mutual information (CMI) and outperforms other methods including reinforcement learning methods.

While there are many methods for DFS, these studies have assumed that the set of measurable feature is fixed and they cannot deal with the situation in which the measurable feature set differs for instance to instance.
This issue arises in certain practical scenarios. 
For example, consider a problem where a school needs to check the trustworthiness of a student for their entrance exam based on the ratings of those who know the person well. 
The committee members can interview acquaintances and obtain their ratings of the student as feature values.
In this case, the set of acquaintances of the candidate is different for each person, so the committee has to deal with an arbitrary number of features for its screening.
This study addresses this problem, namely, DFS with variable feature sets.

Additionally, we propose a novel concept, ''features of features'', which are the prior information about each feature, to identify what each feature is under the problem of variable feature sets. 
By introducing them, we can treat the variable feature sets with which we can not represent features in the conventional manner.
While similar concepts are introduced in some other studies \cite{features_of_features,side_information,property_selection}, No other prior study has introduced them for DFS, to the best of our knowledge.

In this paper, we propose a deep learning method that employs features of features for DFS with variable feature sets.
Our method builds upon the existing DFS method \cite{cmi_feature} by incorporating features of features.
We conducted experiments on a number of datasets to evaluate the performance of our method under the setting of variable feature sets. 
We compare our proposed method with both the existing method and the random feature selection model. The results demonstrated that our method was able to identify more useful features and achieve more accurate classification by leveraging features of features, even if the feature set differed between instances. 

We summarize the main contributions of this study as follows:
\begin{itemize}
\item We address the novel problem of dynamic selecting features where the set of measurable features varies from instance to instance.
\item We propose a model that leverages features of features to tackle this new dynamic feature selection (DFS) problem.
\item We experimentally validate the effectiveness of our proposed method on image and document classification datasets.
\end{itemize}

\section{Related Work}
\label{relates}

DFS is the problem of feature selection, which is different from traditional static feature selection \cite{static_feature_selection}.
While typical feature selection methods are to select only one feature set for all instances in the whole dataset, we flexibly select features one by one based on the value of selected features in DFS.
DFS obviously includes such typical feature selection as a subclass and it is a more natural setting in some situations in which human experts solve problems by gathering information ad hoc.
Originally, it was studied in \cite{active_feature_acquisition} with the name of active feature acquisition.

DFS has often been addressed by using reinforcement learning, where the state is represented as the selected and revealed feature values and the agent selects features one by one \cite{datum-wise,RL_selection,joint_afa}.
In these methods, the agent gets the reward based on some criteria e.g. acquisition cost and loss of prediction, certainly of prediction \cite{opportunistic_learning} or uncertainly of feature value \cite{surrogate_models}.

Various approaches besides reinforcement learning have also been explored in prior studies.
Some methods are based on generative models \cite{EDDI,information_persuit}, active learning \cite{active_selection}, or other statistics \cite{question_ordering,matrix_completion_afa}.
Besides these approaches, recent researches directly train deep neural networks for the loss function of DFS \cite{difa,cmi_feature}, which has a high scope for extension.

However, above all the methods assume that the feature set that we can select is fixed, and they did not consider the situation in which the available feature set is various or the unseen feature appears in the inference time.
Therefore, in this study, we define a new problem of DFS with various feature sets, and we propose a deep learning method, which extends the existing method for DFS with ''features of features'' as prior information.
To the best of our knowledge, though several studies have incorporated prior information for DFS \cite{estimate_cmi,contextual_selection}, no other study has incorporated prior information for each feature similar to our features of features.

Shortly, we assume that features of features as the numerical description of each feature such as natural language (using word2vec, doc2vec), censor location, or other properties.
While features of features we introduce have been studied in a few studies, there are a few studies that investigate a similar concept.
One study introduces features of features to compute regularization term for elastic net \cite{features_of_features}, and the other incorporates ''side-information'' to improve the prediction performance \cite{side_information}.
Similarly, one of the relatively old studies introduces ''property'', which is computed from entire feature values for feature selection \cite{property_selection}.

\section{Problem Definition}
\label{problem}

In this section, we introduce features of features, describe the problem of DFS with variable feature sets, and define the notation used throughout the paper.

\subsection{Variable Feature Set}
\label{prob_vfs}

We consider the supervised learning problem that the feature sets are variable.
Let $x$ denote the input features and $y$ a response variable, and we denote the dataset with $N$ instances as $\mathcal{D}=\{x^{(n)},y^{(n)}\}_{n=1}^{N}$.
We now assume the variable feature set, or the features in each instance $x^{(n)}$ has a different set of features.
Formally, we define it using an index set.
Assuming that there is an entire index set $\Lambda$ which is a finite or infinite set, each instance takes a unique finite subset of the entire index set $\Lambda^{(n)} \subset \Lambda$, which indicates the contents of $x^{(n)}$.
Namely, $\Lambda^{(n)}=\{\lambda^{(n)}_1,...,\lambda^{(n)}_{d^{(n)}}\}$ and $x^{(n)}=\{x_\lambda:\lambda \in \Lambda^{(n)}\}$.

Note that this problem includes the typical supervised learning problem, in which the feature sets are fixed. 
In typical supervised learning, all instances have the same feature set $x^{(n)}=\{x_1^{(n)},..., x^{(n)}_d\}$.
This is the case that the entire index set $\Lambda=\{1,...,d\}$ and the subset of the index set is always the same, that is $\Lambda^{(n)}=\Lambda$.

\subsection{Features of Features}
\label{prob_fof}

We newly introduce features of features to identify each feature, because, under the assumptions of variable feature sets, we cannot identify each feature as the element of a vector.
We define features of features as $z_\lambda \in \mathbb{R}^c$ which is corresponding to $x_\lambda$, and we assume that we can access $z_\lambda$ for any $\lambda \in \Lambda$.
We also assume that these features of features $z_\lambda$ characterize the distribution of $x_\lambda$.

\begin{figure}[ht]
   \begin{center}
    \begin{subfigure}{0.27\linewidth}
        \centering
        \includegraphics[scale=0.29]{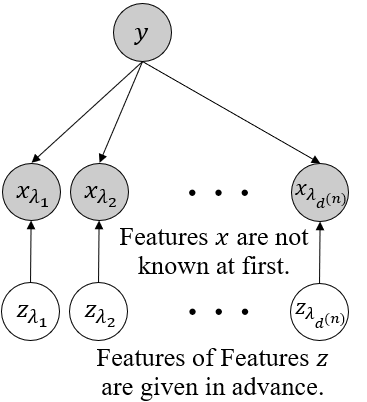}
        \caption{Graphical Model}
        \label{graphical}
    \end{subfigure}
    \begin{subfigure}{0.41\linewidth}
        \centering
        \includegraphics[scale=0.29]{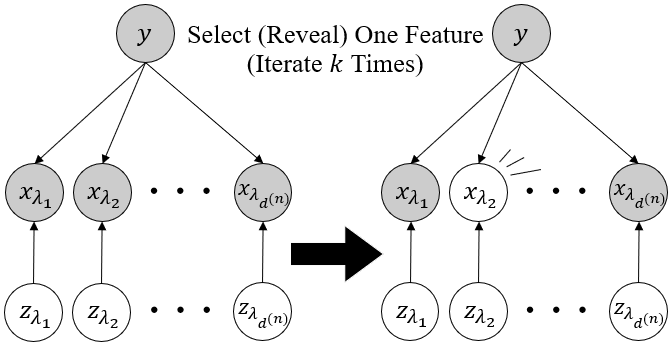}
        \caption{Selection}
        \label{fig:selection}
    \end{subfigure}
    \begin{subfigure}{0.29\linewidth}
        \centering
        \includegraphics[scale=0.29]{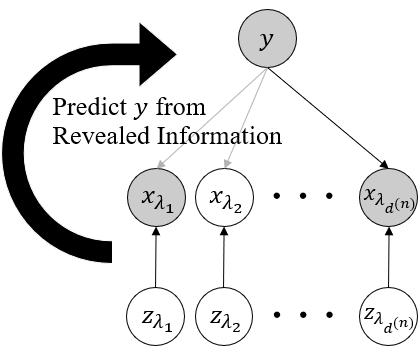}
        \caption{Prediction}
        \label{fig:prediction}
    \end{subfigure}
   \end{center}
   \caption{Graphical model of our problem setting and selection/prediction process: 
            $y$ is the response variable, $x_{\lambda_i}$ is the feature value and $z_{\lambda_i}$ is the features of features. 
            The index set $\Lambda^{(n)}=\{\lambda_1,...,\lambda_{d^{(n)}}\}$ can be different from each instance.
            The dependencies between $x_{\lambda_i}$ and $y$ follow the assumption of the naive Bayesian model.
            The value of $x_{\lambda_i}$ depends on $y$ and $z_{\lambda_i}$.
            Variables with a gray background are unknown and those with a white background are known.}
   \label{graphical_model}
\end{figure}

We assume the dependencies of the naive Bayesian model for $x^{(n)}$ and $y^{(n)}$ following the prior study \cite{greedy_policy,cmi_feature}, and the overall variable dependencies are explained as the graphical model in \cref{graphical}.
From the assumption of the naive Bayesian and of the property of features of features explained above, we can consider the distribution of each $x_\lambda$ is determined only by $y$ and $z_\lambda$.

In conclusion, the entire dataset we consider is the tuple with three elements; $\mathcal{D}=\{x^{(n)},z^{(n)},y^{(n)}\}_{n=1}^{N}$, where $z^{(n)}=\{z_\lambda:\lambda \in \Lambda^{(n)}\}$.

\subsection{Dynamic Feature Selection}
\label{prob_dfs}

The goal of DFS is to select features that are more helpful for making more accurate predictions, considering the cost of feature value acquisition.
In this study, only consider the number of selected features, that is, we assume the uniform cost for all features.
Let $s^{(n)} \subset \Lambda^{(n)}$ be the subset of the index and $x^{(n)}_s=\{x_\lambda : \lambda \in s^{(n)}\} \subset x^{(n)}$ be the subset of features, which represents the selected and revealed features. 
Our task is to select features $s^{(n)}$ and make a final prediction about $y^{(n)}$ while $s^{(n)}$ is different for each instance.

Our purpose is to find two functions, implemented as neural networks; the policy to select features and the predictor to make predictions.
The policy takes the revealed features $x^{(n)}_s$ and the whole features of features $z^{(n)}$ as input. 
Then, it returns one index and the corresponding feature will be selected and revealed.
We denote the policy as $\pi(x^{(n)}_s, z^{(n)}) \in \Lambda^{(n)}$.
The predictor also takes $x^{(n)}_s$ and $z^{(n)}$ as input and it returns the prediction about $y^{(n)}$.
We denote the prediction and the predictor as $\hat y^{(n)} = f(x_s^{(n)},z^{(n)})$

We innovate a way to select features one by one with the policy until the predetermined number of features is selected.
Firstly, we determine the budget $k$, which is the number of selected features in the end.
In the initial state, we suppose $s^{(n)}=\phi$ and $x^{(n)}_s=\phi$, that is, any features has not revealed. 
Note that, we assume that features of features $z$ are fully revealed in the initial state and they are given as prior information.
This assumption is natural because we consider features of features as the description or the property of each feature on the natural language or numerical expression and they are usually given before the feature values are revealed.
Then, we select a feature with the policy $\pi(x_s,z)$ and reveal the selected feature.
(In other words, $s \leftarrow s \cup \pi(x_s,z)$ and the size of $x_s$ increases by one.)
We repeat this process for $k$ times and make the final prediction with the predictor $f(x_s,z)$ from $|x_s|=k$ features. 
This procedure is shown in \cref{fig:selection,fig:prediction}.

Our ultimate goal is to minimize the loss of the prediction made by the predictor after selecting features by the policy.
Given the loss function $l(\hat y, y)$, the objective function of our problem is
\begin{align}
   \min_{\theta,\phi} \mathbb{E}_{\mathbf{y}, \mathbf{x}} l(f_\theta(\mathbf{x}_s,z),\mathbf{y}),
\end{align}
where $\theta$ and $\phi$ are parameters for parameterized predictor and policy respectively. In addition, $s$ ($|s|=k$) is constructed by the iterative selection of the policy $\pi_\phi(x_s,z)$.

\section{Greedy Dynamic Feature Selection}
\label{background}

This section explains the existing DFS method based on conditional mutual information (CMI) \cite{cmi_feature}. 
Note that this method handles the special case of our problem definition that the feature set is fixed.
Namely, $\Lambda=[d]$ and $\Lambda^{(n)}=\Lambda$ for all $n=1,...,N$.
Therefore, in this section, we note $x=\{x_1,...,x_d\}$ and $x_s=\{x_i : i \in s \subset [d]\}$.
Also, the policy $\pi(x_s)$ and the predictor $f(x_s)$ take only revealed features $x_s$ as input.

\subsection{Greedy Dynamic Feature Selection}
\label{exist_method}

This method aims to find a greedy policy based on CMI and the predictor based on the naive Bayesian, implemented as neural networks such as MLP.
Concretely, the optimal policy $\pi^*$ and predictor $f^*$ have been defined as
\begin{align}
   \label{exist_policy}
   \pi^*(x_s) &= \operatorname{argmax}_{i} I(\mathbf{y};\mathbf{x}_i|x_s), \\
   \label{exist_predictor}
   f^*(x_s) &= p(\mathbf{y}|x_s).
\end{align}

To find optimal policy and predictor, they train two neural networks, $\pi_\phi(x_s) \in [d]$ and $f_\theta(x_s) \in \Delta^K$, by minimizing the following one-step-ahead loss function.
\begin{equation}
   \label{exist_loss}
   \mathbb{E}_{\mathbf{y}, \mathbf{x}_i|x_s}[l(f_\theta(x_s \cup \mathbf{x}_i),\mathbf{y})],
\end{equation}
where $i=\pi_\theta(x_s)$ is the selected index by the policy.

To optimize the objective function, they employed the amortized optimization approach using the reparameterization trick. 
They reformulate the policy $\pi_\phi(x_s) \in \Delta^d$ as the distribution on the index set, and the loss function \cref{exist_loss} can be rewritten as
\begin{equation}
   \label{exist_loss_rewrite}
   \mathbb{E}_{p(x,y)} \mathbb{E}_{p(s)} [\mathbb{E}_{i\sim \pi(x_s)}[l(f(x_s\cup \mathbf{x}_i),\mathbf{y})]].
\end{equation}
While the sampling of index $i \sim \pi(x_s)$ is naively represented as one-hot vector $m \in \{0,1\}^d$, they employ the continuous relaxation of $m \in \Delta^d$ with Concrete distribution \cite{concrete}.
Then, they compute the gradient of the policy $\pi_\phi$ using the reparameterization trick, or input element-wise product $x \odot m$ to the predictor.

In practice, the models are trained by selecting one feature at a time and computing the prediction loss each time. 

The correctness of their approach is theoretically guaranteed by the following theorem.
(see proofs in Appendix of the original paper \cite{cmi_feature}.)

\begin{theorem}
\label{exist_theorem}
    When $l$ is cross-entropy loss, the global optimum of \cref{exist_loss_rewrite} satisfy that the predictor be the Bayes classifier $f_{\theta^*}(x_s)=p(\mathbf{y}|x_s)$ and the predictor put all probability mass on $i^*= \operatorname{argmax}_{i} I(\mathbf{y};\mathbf{x}_i|x_s)$.
\end{theorem}

Based on this theorem, we can train the optimal policy \cref{exist_policy} and predictor \cref{exist_predictor} while minimizing the objective function \cref{exist_loss_rewrite}.

\section{Proposed Method}
\label{proposed}

To deal with the problem in \cref{problem}, we describe our proposed method for the DFS problem of variable feature sets with features of features in this section.

\subsection{Dynamic Feature Selection with Features of Features}
\label{proposed_method}

We aim to find the optimal policy and the predictor defined similarly as \cref{exist_method} with features of features.
\begin{align}
   \label{proposed_policy}
   \pi^*(x_s,z) &= \operatorname{argmax}_{i} I(\mathbf{y};\mathbf{x}_i|x_s,z), \\
   \label{proposed_predictor}
   f^*(x_s,z) &= p(\mathbf{y}|x_s,z).
\end{align}

We simply introduce features of features for the DFS model explained in \cref{exist_method}.
The predictor is represented as parameterized neural networks $f_\theta(x_s,z) \in \Delta^{d^{(n)}}$.
According to the amortized optimization method, we use the policy of distribution version $\pi_\phi(x_s,z) \in \Delta^{(d^{(n)})}$.

We define the loss function with features of features, referencing \cref{exist_loss_rewrite}.
\begin{equation}
   \label{proposed_loss}
   \mathbb{E}_{p(x,y,z)} \mathbb{E}_{p(s)} [\mathbb{E}_{i\sim \pi(x_s,z)}[l(f(x_s\cup \mathbf{x}_i,z),\mathbf{y})]].
\end{equation}

As the \cref{exist_method}, we also use Concrete Distribution and reparameterization tricks to compute the gradient of the policy.
The algorithm for training our model is shown in Algorithm \ref{our_algorithm}.

\begin{figure}[!t]
   \begin{algorithm}[H]
       \caption{Training Algorithm with Features of Features}
       \label{our_algorithm}
       \begin{algorithmic}
           \renewcommand{\algorithmicrequire}{\textbf{Input:}}
           \renewcommand{\algorithmicensure}{\textbf{Output:}}
           \REQUIRE features $x$,  features of features $Z$, ground truth $y$, the number of feature selection $k$, temperature for Concrete Distribution $T=\{\tau_1,...,\tau_n\}$
           \ENSURE  classifier $f(x,Z;\theta)$, policy $\pi(x,Z;\phi)$
           \STATE initialize $f(x,Z;\theta),\pi(x,Z;\phi)$
           \WHILE{not convergence}
           \STATE retrieve $x$, $Z$, $y$ from training data
           \STATE initialize $\mathcal{L}=0$, $m=[0,...,0]$
           \FOR{$j=1$ to $k$}
           \STATE $\alpha = \pi(x\odot m;\phi)$, 
           \STATE // Sample $G_i$ from Gumbel distribution (concrete distribution)
           \STATE $\tilde{m}=\mathrm{max}(m,\mathrm{softmax}(G+\alpha,\tau))$ \quad
           \STATE $m=\mathrm{max}(m,\mathrm{softmax}(G+\alpha,0))$
           \STATE $\mathcal{L}\longleftarrow\mathcal{L}+l(f(x\odot \tilde{m},Z;\theta),y)$
           \ENDFOR
           \STATE$\phi \longleftarrow \phi - \gamma \nabla_{\phi}\mathcal{L}$, 
           $\theta \longleftarrow \theta - \gamma \nabla_{\theta}\mathcal{L}$
           \ENDWHILE
       \end{algorithmic}
\end{algorithm}
\end{figure}

\begin{figure}[ht]
   \begin{center}
      \includegraphics[scale=0.31]{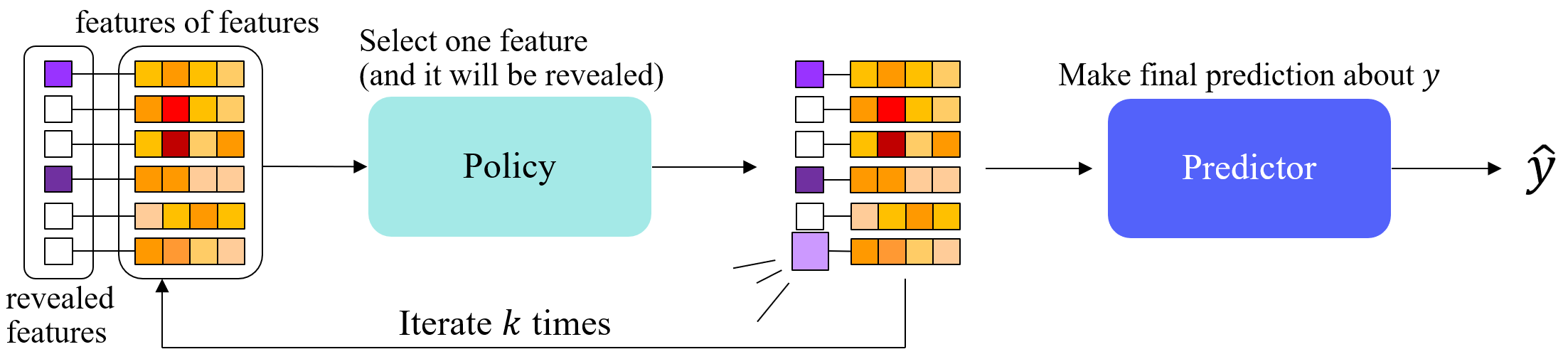}
   \end{center}
   \caption{Our model and data flow. First, the policy selects features iteratively based on revealed features and features of features, then the predictor makes the final prediction from revealed information.}
   \label{data_flow}
\end{figure}

The input of our model consists of not only feature values (and mask) but features of features, that cannot be represented as a 1-D vector.
So, we simply concatenate the feature value and corresponding features of features.
A feature vector with index $\lambda$ is represented as $[x_\lambda,m_\lambda,z_\lambda]$, and the input of instance $n$ is $\{[x_\lambda^{(n)},m^{(n)}_\lambda,z^{(n)}_\lambda]:\lambda \in \Lambda^{(n)}\}$.
Inputting these feature vectors, the neural networks for the policy and predictor should not be affected by the order of feature vectors, or our policy and predictor must have equivariance and invariance for the permutation of feature vectors. 
Therefore, as the model that satisfies these properties, we use equivariant or invariant layers for permutation of input, such as attention mechanism \cite{attention_is_all_you_need}. We implement the policy and predictor with these layers based on the concept of DeepSets \cite{deepsets} or FT-Transformer \cite{deep_tabular_data}.
The model and data flow are shown in \cref{data_flow}.

The theoretical explanation of our DFS model with features of features can be described as the following theorem.

\begin{theorem}
\label{proposed_theorem}
When $l$ is cross-entropy loss, the global optimum of \cref{proposed_loss} satisfy that the predictor be the Bayes classifier $f_{\theta^*}(x_s,z)=p(\mathbf{y}|x_s,z)$ and the policy put all probability mass on $i^*= \operatorname{argmax}_{i} I(\mathbf{y};\mathbf{x}_i|x_s,z)$.
\end{theorem}

\cref{proposed_theorem} is the augmented version of the \cref{exist_theorem} for the feature of features, and it can be proven with the conditional independence of $z_\lambda$.
From this, we can say that the minimization of \cref{proposed_loss} is equivalent to finding the optimal policy and predictor defined in \cref{proposed_policy} and \cref{proposed_predictor}.

\section{Experiments}
\label{experiment}

We experimentally confirmed the utility of our proposed method on image and document datasets. 
To simulate the situation of variable feature sets, we restricted the available features for these datasets. 
In each experiment, we compared three models: our proposed method with features of features, our method in which the selector selects features randomly as an ablation study, and the base existing method \cite{cmi_feature}, which is without features of features.

\subsection{Image Classification}
\label{image}

The datasets that we used MNIST, FashionMNIST, and CIFAR10, which are image classification datasets of 10 classes. 
In these datasets, the features to be selected are pixel values.
We also used the position of each feature as features of features. We normalized them into the range of $[0,1]$. 
In CIFAR-10, because each pixel is represented in three channels, we treated a tuple of three values as one feature. 

To reproduce the variable feature sets, we sampled 100 available features for each image and the models select features from them. 

\begin{figure}[tb]
    \centering
    \begin{subfigure}{0.32\linewidth}
        \centering
        \includegraphics[scale=0.25]{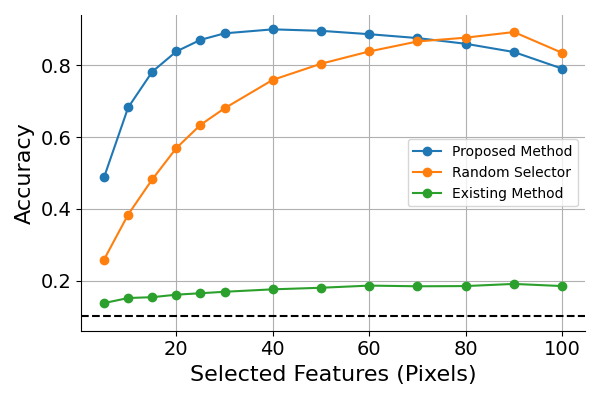}
        \caption{MNIST}
        \label{mnist acc}
    \end{subfigure}
    \begin{subfigure}{0.32\linewidth}
        \centering
        \includegraphics[scale=0.25]{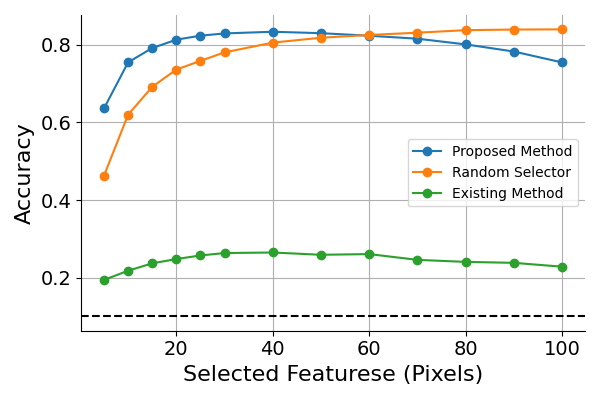}
        \caption{FashionMNIST}
        \label{fashionmnist acc}
    \end{subfigure}
    \begin{subfigure}{0.32\linewidth}
        \centering
        \includegraphics[scale=0.25]{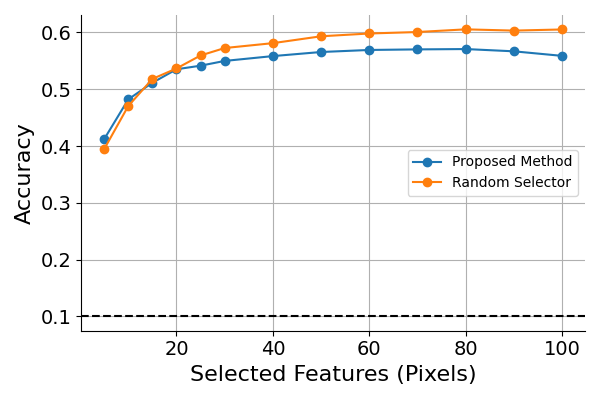}
        \caption{CIFAR-10}
        \label{cifar-10 acc}
    \end{subfigure}
    \caption{Accuracy for image classification. The horizontal axis is the number of selected features, and the vertical axis is the accuracy of classification. The dotted line in each figure is the chance rate.}
    \label{fig:classification_accuracy}
\end{figure}

\begin{figure}
    \begin{subfigure}{0.32\linewidth}
      \centering
      \includegraphics[scale=0.18]{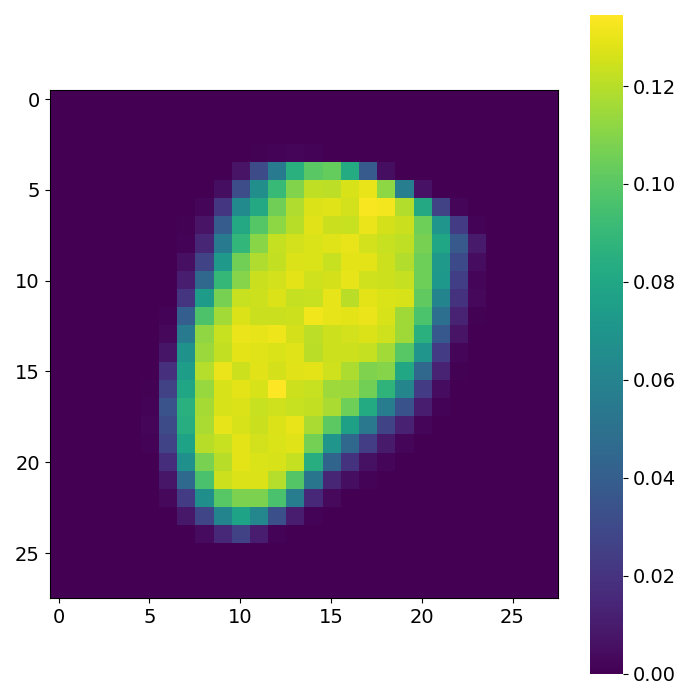}
      \caption{MNIST}
    \end{subfigure}
    \begin{subfigure}{0.32\linewidth}
      \centering
      \includegraphics[scale=0.18]{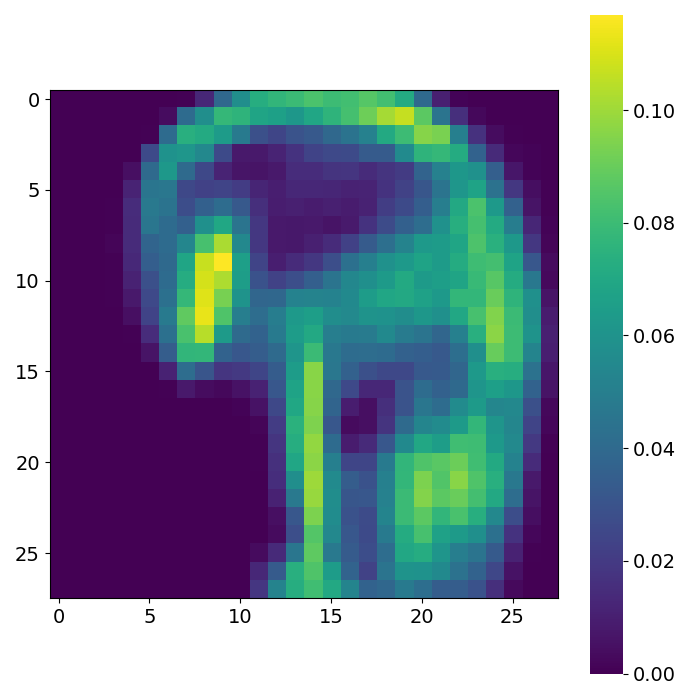}
      \caption{FashionMNIST}
    \end{subfigure}
    \begin{subfigure}{0.32\linewidth}
      \centering
      \includegraphics[scale=0.18]{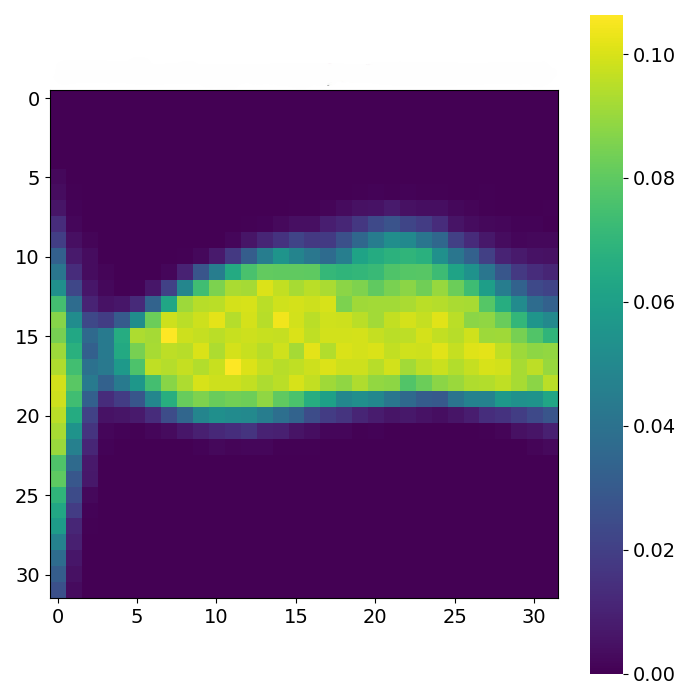}
      \caption{CIFAR-10}
    \end{subfigure}
    \caption{Frequency of feature selection for each image dataset. More frequently selected positions are lighter in color.}
    \label{image freq}
\end{figure}

\paragraph{Result.} The results are shown in \cref{fig:classification_accuracy} 
These figures present the accuracy of classification for each feature budget. 
The existing method could not perform well, because the sets of available features are different from each instance, and the correspondence between the feature and the dimension of the input vector is lost. 
Meanwhile, the proposed method utilizes features of features that identify the position of each feature and show reasonable performance. 
The proposed method outperformed the model with a random selector when the feature budget is small, which shows that the selector selected useful features for classification.

The frequency of selection for each pixel is provided as \cref{image freq}. 
In MNIST, pixels in the center are more selected, while in FashionMNIST, pixels in the surrounding area were more selected than those in the center, which implies that the selector could select features adopting specific tasks.

\subsection{Document Classification}
\label{document}

We use two datasets; BBCSport and 20NEWS, which are the classifications of articles by their topics. In BBCSport, the classes are five topics. In 20NEWS, the classes are 20 topics.

Each article is represented as bag-of-words and the features are the number of appearances of each word.
We set features of features as the vector representation of each word with word2vec \cite{word2vec}. 
To compute vector representation, we used a pre-trained word2vec model (GoogleNEWS-vectors-negative300.bin). 
Because some words are not included in the vocabulary of the word2vec model, we excluded such words in advance.

As with the image dataset, we sampled 2000 available features for each article. 
As the structure of the predictor, we used MLP that takes as input the mean of embedding vectors of words that were revealed to appear in the article.

\begin{figure}[tb]
    \begin{subfigure}{0.5\linewidth}
        \centering
        \includegraphics[scale=0.35]{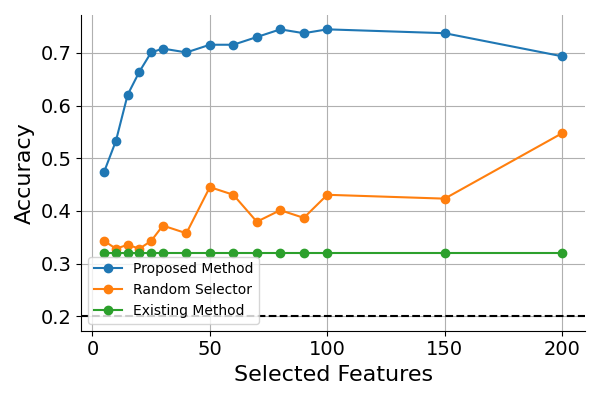}
        \subcaption{BBCSport}
        \label{bbcsport}
    \end{subfigure}
    \begin{subfigure}{0.5\linewidth}
        \centering
        \includegraphics[scale=0.35]{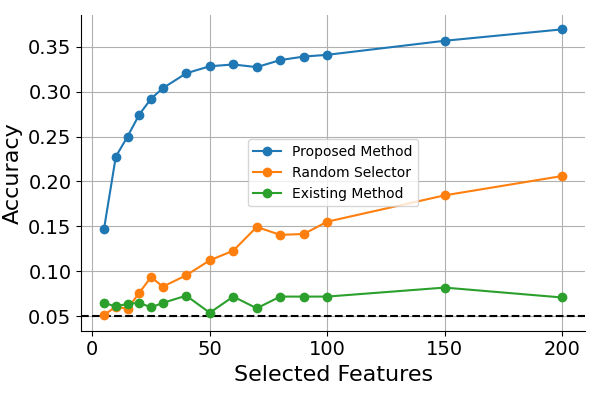}
        \subcaption{20NEWS}
        \label{20NEWS}
    \end{subfigure}
    \caption{The trade-off of accuracy and feature budget for document classification. The dotted line in each figure is the chance rate.}
    \label{fig:document}
\end{figure}

\paragraph{Result.} The results are shown in \cref{fig:document}. 
Similar to the experiments on images, while the existing method could not provide valid classification, our proposed method could perform better using vector representation of words as features of features. 
Furthermore, from the comparison of the proposed method and the model with a random selector, both predictor and selector could treat features of features well.

\section{Discussion}

Though we experimented on image classification and document classification datasets, the setting of our experiments was the simulation of variable feature sets. 
It will be necessary to conduct the experiments according to actual scenarios, such as the verification of trustworthiness based on the acquaintances described in \cref{intro}.

About the features of features, we confirmed the utility of the position of pixel and vector embedding of words. Furthermore, if the embedding of a short document can be used as features of features, it will bring a wider range of applications of our method because many datasets have descriptions in natural language for each feature. 
More generally, employing representation learning methods to get embeddings of the concept about each feature may be used as features of features and bring more applications.

As the technical limitation, in our implementation, to aggregate the information of revealed features, the attention mechanism was used in the selector (and predictor in some tasks).
Therefore, the computational complexity was proportional to the square of the input size, and this was the main bottleneck. This limitation may be solved by replacing the self-attention layer with more efficient neural network layers.

\section{Conclusion}
\label{concl}

To deal with the case where the set of features that can be selected varies from instance to instance, we proposed a method to perform DFS by providing prior information about each feature as features of features. 
The proposed method trains two networks: a policy and a classifier. 
The policy takes as input the values of features that have already been measured and features of features and selects the next feature to be measured. 
After the policy selects a certain number of features, the classifier performs a final classification based on the measured feature values and features of features. 
To deal with the problem of variable feature sets and make use of features of features, the policy and classifier are networks with the equivariance and invariance for permutation, respectively.

Experiments on the image and document classification datasets showed that the proposed method can select useful features for classification based on features of features and perform appropriate classification even when the set of features varies from instance to instance.

\begin{credits}
\subsubsection{\ackname} I would like to express my sincere gratitude to the members of our laboratory for providing us with a great deal of knowledge and advice through daily discussions and for their help in paper writing.

\end{credits}
\bibliographystyle{splncs04}
\bibliography{main}
\end{document}